\documentclass[sigconf]{acmart}

\AtBeginDocument{%
  \providecommand\BibTeX{{%
    \normalfont B\kern-0.5em{\scshape i\kern-0.25em b}\kern-0.8em\TeX}}}

\setcopyright{none}
\renewcommand\footnotetextcopyrightpermission[1]{} % removes footnote with conference information in first column
\copyrightyear{2020}
\acmYear{2020}
\setcopyright{acmcopyright}\acmConference[MM '20]{Proceedings of the 28th ACM International Conference on Multimedia}{October 12--16, 2020}{Seattle, WA, USA}
\acmBooktitle{Proceedings of the 28th ACM International Conference on Multimedia (MM '20), October 12--16, 2020, Seattle, WA, USA}
\acmPrice{15.00}
\acmDOI{10.1145/3394171.3413767}
\acmISBN{978-1-4503-7988-5/20/10}

\newenvironment{myitemize}{\begin{list}{$\bullet$}
		{\setlength{\topsep}{1mm}
			\setlength{\itemsep}{0.25mm}
			\setlength{\parsep}{0.25mm}
			\setlength{\itemindent}{0mm}
			\setlength{\partopsep}{0mm}
			\setlength{\labelwidth}{15mm}
			\setlength{\leftmargin}{4mm}}}{\end{list}}
\usepackage{algorithm}
\usepackage[noend]{algorithmic}
\usepackage{subcaption}
\usepackage{xcolor}
\usepackage{hyperref}

\begin{document}
\fancyhead{}
\title{Distributed Multi-agent Video Fast-forwarding}

\author{Shuyue Lan}
\affiliation{%
  \institution{Northwestern University}
%   \city{Evanston}
%   \state{IL}
}
\email{shuyuelan2018@u.northwestern.edu}

\author{Zhilu Wang}
\affiliation{%
  \institution{Northwestern University}
%   \city{Evanston}
%   \state{IL}
}
\email{zhiluwang2018@u.northwestern.edu}

\author{Amit K. Roy-Chowdhury}
\affiliation{%
  \institution{University of California, Riverside}
%   \city{Riverside}
%   \state{CA}
}
\email{amitrc@ece.ucr.edu}

\author{Ermin Wei}
\affiliation{%
  \institution{Northwestern University}
%   \city{Evanston}
%   \state{IL}
}
\email{ermin.wei@northwestern.edu}

\author{Qi Zhu}
\affiliation{%
  \institution{Northwestern University}
%   \city{Evanston}
%   \state{IL}
}
\email{qzhu@northwestern.edu}

\renewcommand{\shortauthors}{Lan and Wang, et al.}

\begin{abstract}
 In many intelligent systems, a network of agents collaboratively perceives the environment for better and more efficient situation awareness. As these agents often have limited resources, it could be greatly beneficial to identify the content overlapping among camera views from different agents and leverage it for reducing the processing, transmission and storage of redundant/unimportant video frames.  
This paper presents a consensus-based distributed multi-agent video fast-forwarding framework, named \textbf{DMVF}, that fast-forwards multi-view video streams collaboratively and adaptively. In our framework, each camera view is addressed by a reinforcement learning based fast-forwarding agent, which periodically chooses from multiple strategies to selectively process video frames and transmits the selected frames at adjustable paces. 
During every adaptation period, each agent communicates with a number of neighboring agents, evaluates the importance of the selected frames from itself and those from its neighbors, refines such evaluation together with other agents via a system-wide consensus algorithm, and uses such evaluation to decide their strategy for the next period. Compared with approaches in the literature on a real-world surveillance video dataset VideoWeb, our method significantly improves the coverage of important frames and also reduces the number of frames processed in the system.

\end{abstract}




\maketitle

\section{Introduction}

In many intelligent multi-agent systems, a network of agents with cameras can jointly perform tasks such as search and rescue, surveillance, and environment monitoring. These cameras may be fixed surveillance cameras, or built-in cameras on robots/drones. They generate a large amount of videos from different viewpoints, and sometimes transmit these videos to a cloud server for further analysis, decision making, and storage. 
For many applications, the processing and transmission of the video frames from the agents to the server needs to be performed in an online manner at real time or near real time. However, the agents often have limited computation, communication, storage, and energy resources~\cite{akyildiz2007survey,ma2013survey,singh2014survey}, which make it challenging or even intractable to process and transmit all the video data at real time.  
Thus, methods that can select an \emph{informative subset} of the video frames for processing, transmission and storage are greatly needed to reduce the resource requirements. 

In the literature,  video summarization techniques, which generate a compact summary of the original video, have been widely studied~\cite{elhamifar2017online,gygli2015video,panda2017weakly,zhang2016summary,zhang2016video,zhao2014quasi}. Multi-view video summarization that handles video streams from multiple cameras has also been addressed in several works~\cite{fu2010multi,panda2016video,panda2017multi,elfeki2018multi,ou2015line}. However, these methods need to process an entire video (i.e., every frame in the video) and often take a long time for generating a summary, which are not suitable for real-time and online applications. Some methods~\cite{cheng2009smartplayer,halperin2017egosampling,joshi2015real,petrovic2005adaptive,poleg2015egosampling,ramos2016fast,silva2016towards} have been developed for video fast-forwarding by adjusting the playback speed of a video, but they still require processing of the entire video and will not reduce the amount of data to be transmitted when used in a multi-agent scenario. 

A recent work~\cite{lan2018ffnet} performs fast-forwarding for a single agent in an online manner, and processes a fraction of the video frames by automatically skipping unimportant frames via reinforcement learning. However, the method considers only a single agent and cannot be easily extended to the multi-agent domain. Moreover, as we observe that there are often significant overlaps among the videos captured by the different agents, we start by asking the following question: \emph{Is it possible to develop a method for multiple agents to collaboratively perform fast-forwarding that is efficient, causal, online and results in an informative summary for the scene?}

\begin{figure}[t]
	\begin{center}
		\includegraphics[width=0.77\linewidth]{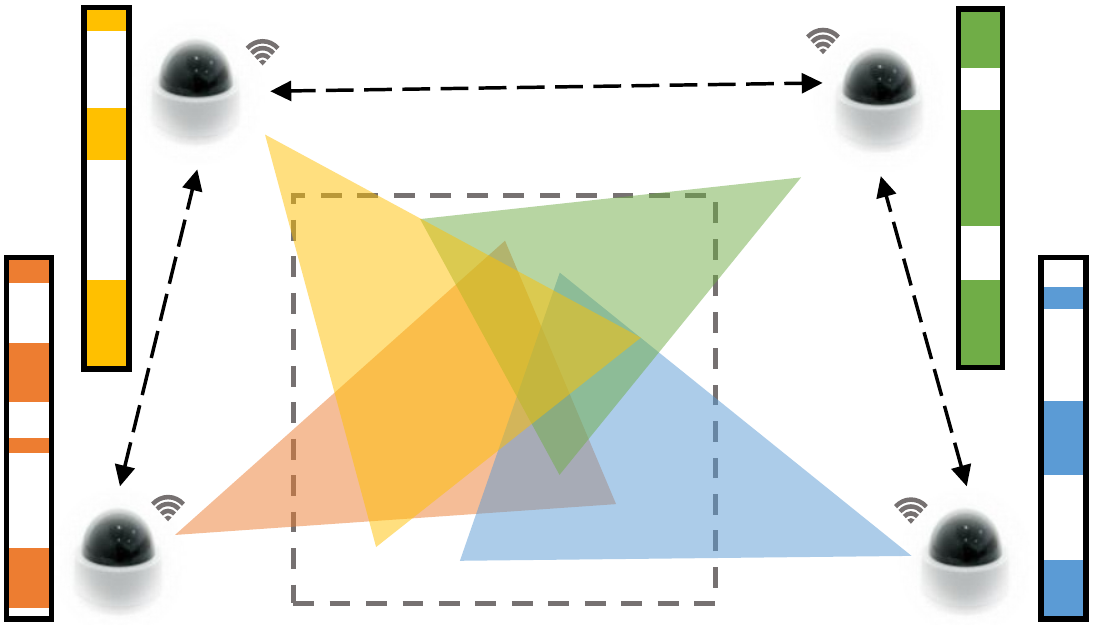}
	\end{center}
	\caption{\textbf{Illustration of distributed multi-agent video fast-forwarding setup}. Multiple cameras at different locations observe the same environment from different points of view. Each view has a fast-forwarding agent that periodically adapts its strategy for selectively processing input video frames. At every adaptation period, agents communicate with each other to collaboratively 
	decide each agent's fast-forwarding strategy for the next period. 
	}
	\label{overview}
\end{figure}

In this paper, we introduce the \textbf{D}istributed \textbf{M}ulti-agent \textbf{V}ideo \textbf{F}ast-Forwarding framework (\textbf{DMVF}), a consensus-based  framework that collaboratively fast-forwards videos at different views for efficient processing, transmission and storage of video data (see Fig.~\ref{overview}).
In our target scenario, cameras at multiple locations observe the same environment from different views that may be overlapping. Each camera embeds a reinforcement learning based fast-forwarding agent with multiple strategies to choose from, i.e., it can skip the frames of its video input at different paces (e.g., slow, normal or fast). Agents are connected by a predetermined undirected communication network, where each agent can communicate with a set of neighboring agents. We also assume the communication graph is connected.

During operation, each agent adapts its fast-forwarding strategy periodically based on how important its selected frames are when compared with other agents' frames.  
At every adaptation period, each agent evaluates the importance of the selected frames from itself and those from its neighbors by comparing their similarities. Intuitively, if an agent's frames can better represent/cover the views of other agents, they are regarded as more important.
The agents then refine their evaluation by running a system-wide consensus algorithm~\cite{tsitsiklis1984problems} and reach an agreement on the importance score for every agent's selected frames.
Based on the score ranking and the system requirement, each agent selects a fast-forward strategy for its next adaptation period. Intuitively, agents with lower scores on their selected frames could be given a faster pace for the next period to reduce their processing and transmission load, while the ones with higher scores should be given the same (or slower) pace.

It is worth noting that in our approach, each agent only processes a very small portion of the frames, which greatly helps reduce the computation load on resource-limited embedded platforms. Agents also do not transmit their entire video streams but only a fraction of them. From the system perspective, both intra-view redundancy at each agent and the inter-view redundancy across different agents are reduced. Furthermore, the online and causal nature of our proposed DMVF framework enables the users to begin fast-forwarding at any point when executing certain multi-agent perception tasks. Our approach is particularly useful for resource-constrained and time-critical systems such as multi-robot systems. 

To summarize, the following are the main contributions.
\begin{myitemize}
	\item We formulate the multi-agent video fast-forwarding problem as a multi-agent reinforcement learning problem. Each agent can fast-forward its video input without processing the entire video.
	\item We design a distributed and consensus-based framework that enables adaptive fast-forwarding strategies/paces for all agents and optimizes the computation and communication load globally.
	\item We demonstrate the effectiveness of DMVF on a challenging multi-view dataset, VideoWeb~\cite{denina2011videoweb}, achieving real-time speed on a practical embedded platform. Compared with other methods in the literature, our approach achieves significantly better coverage of the important frames/events and reduces the computation and storage load, as well as the communication to the cloud.
\end{myitemize}
\vspace{-6pt}

\section{Related Works}

\subsection{Video Fast-forwarding}
Video fast-forwarding is applied when users are not interested in parts of the video. Some commercial video players offer manual control on the playback speed, e.g., Apple QuickTime Player with 2, 5 and 10 multi-speed fast-forward. Researchers adapt the playback speed based on the motion activity patterns present in a video~\cite{cheng2009smartplayer,peker2003extended,peker2001constant} and the similarity of each candidate clip to the query clip~\cite{petrovic2005adaptive}. Other works focus on developing fast-forwarding policies using mutual information between frames~\cite{jiang2010new,jiang2011smart} and shortest path distance over the semantic graph built from frames~\cite{ramos2016fast,silva2016towards}. Another family of work (hyperlapse)~\cite{poleg2015egosampling,halperin2017egosampling,joshi2015real} fast-forwards videos with the objective of speed-up and smoothing. More recently, Lan et al.~\cite{lan2018ffnet} propose an online deep reinforcement learning agent for skipping frames and fast-forwarding. Different from these methods, our work focuses on multi-agent video fast-forwarding that collaboratively and distributively fast-forwards videos in each view based on the information from its own perception and neighbors'.

\subsection{Video Summarization}
The objective of video summarization is to generate a compact subset of videos that can describe the main content of the original video.  Many offline methods, which require the entire video being available before processing,  are developed with unsupervised learning~\cite{elhamifar2012see,gygli2014creating,Att2005,Top2014,elhamifar2019unsupervised} or supervised learning based on video-summary labels~\cite{gong2014diverse,gygli2015video,zhang2016summary,zhang2016video,Category2014,panda2017weakly,wu2019adaframe}. There also work on summarization for crawled web images/videos~\cite{khosla2013large,Joint2014,song2015tvsum,panda2017collaborative}  and photo albums~\cite{sigurdsson2016learning}. Learning video summarization from unpaired data is studied in~\cite{rochan2019video}. The other branch of work is online video summarization, which summarizes videos by automatically scanning in an online fashion. Various methods are proposed with submodular optimization~\cite{elhamifar2017online}, Gaussian mixture model~\cite{ou2014low}, and online dictionary learning~\cite{zhao2014quasi}.

What is more related to our work is video summarization from multi-view videos. In~\cite{fu2010multi}, the authors introduce the problem of multi-view video summarization and solve it with random walk over spatio-temporal graphs. Joint embedding and sparse optimization are proposed in~\cite{panda2016video,panda2017multi}. More recently, Elfeki et. al~\cite{elfeki2018multi} adapt DPP (Determinantal Point Processes) to multi-view for offline multi-view video summarization. All these methods require the availability of all frames from each view. In~\cite{ou2015line}, the authors propose a two-stage system, i.e., online single-view summarization and distributed view selection for the multi-view case. 
Different from these previous methods, our approach does not process all the frames, which significantly reduces computation and communication load, and it collaboratively fast-forwards the videos of multiple agents to further improve the efficiency and coverage.

\begin{figure*}[t]
	\begin{center}
		\includegraphics[width=0.85\linewidth]{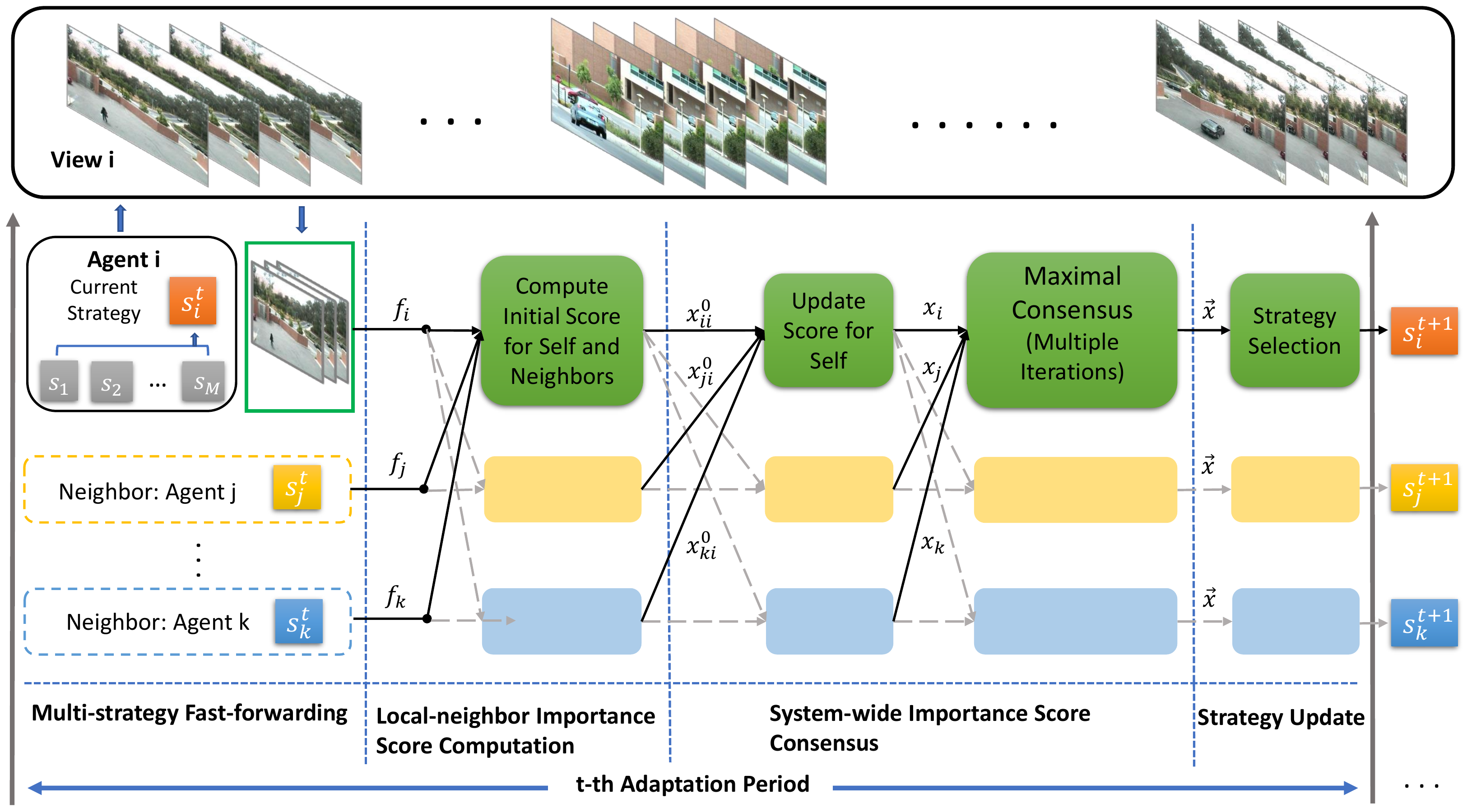}
	\end{center}

	\caption{Our DMVF framework. At every adaptation period $t$, each agent $i$ first fast-forwards its video input with current strategy $s_i^t$ and selects a set of frames $f_i$ (see footnote~\ref{fn:superscript}). It then receives neighbor agents' selected frames (e.g., $f_j$ and $f_k$) and computes an initial importance score for itself and its neighbors. Afterwards, agent $i$ refines and finalizes the importance score with other agents via a system-wide consensus algorithm (maximal consensus is shown in the figure). Based on this importance score vector $\vec{x}$, agent $i$ chooses its strategy for the next period $s_i^{t+1}$ (so does every other agent).} 
	\label{framework}
\end{figure*}
\vspace{-6pt}

\subsection{Distributed Consensus}

A fundamental problem in distributed multi-agent systems is the minimization of a sum of local objective functions while maintaining agreement over the decision variable, often referred to as consensus optimization. Seminal work in~\cite{tsitsiklis1984problems} proposes a distributed consensus protocol for achieving agreement in a multi-agent setting by iteratively taking weighted average with local neighbors. The work in~\cite{nedic2009distributed} presents a distributed gradient descent (DGD) method, where each agent iteratively updates its local estimate of the decision variable by executing a local gradient descent step and a consensus step.  Follow-up works~\cite{JJSubgradient,NOPConstrained,LORandomNetwork,MateiBaras, aliLinMorse, nedic2011asynchronous} extend this method to other settings, include stochastic network, constrained problems, and noisy environment.  
More recently, EXTRA~\cite{shi2015extra}, which takes a careful combination of gradient and consensus steps, is proposed to improve convergence speed and is shown to achieve linear convergence with constant stepsize. In computer vision, consensus based methods are used applications such as human post estimation~\cite{lifshitz2016human}, background subtraction~\cite{wang2006background}, bundle adjustment~\cite{Eriksson_2016_CVPR} and multi-target tracking~\cite{kamal2015distributed}, etc.  
To the best of our knowledge, this is the first distributed consensus based work to address multi-agent video fast-forwarding.

\section{Methodology}

\subsection{Problem and Solution Overview}

Our objective is to collaboratively fast-forward multi-view videos from different agents by adapting the skipping strategy of each agent in an efficient, online and distributed manner. Fig.~\ref{framework} shows the workflow design of our framework (taking one agent $i$ for illustration). Given the incoming multi-view video streams $V = \left\{v_1, \cdots, v_N\right\}$ captured at different agents, our goal is to generate a final summary $F = \left\{f_1,\cdots, f_N\right\}$ for the scene while reducing the computation, communication, and storage load. 

In our framework, the fast-forwarding agent of each view is modeled as a reinforcement learning agent with multiple available strategies $S=\left\{s_m, m = 1, \cdots , M\right\}$. During operation, at every adaptation period $t$ (with the period length as $T$), each agent $i$ fast-forwards its own video stream with a current strategy $s_i^t \in S$ and selects a subset of frames $f_i$~\footnote{\label{fn:superscript}Strictly speaking, $f_i$ should have a superscript $t$ to represent current period (i.e., $f_i^t$). We omit it for better readability wherever it does not cause confusion. The same goes for many other symbols in this section, such as $x_{ii}^0, x_i, \vec{x}$.}. Note that the frames being skipped are not processed nor transmitted. The details of this step are introduced in Section~\ref{sec:agent}. Agent $i$ then communicates with its neighbors and receives their selected frames, e.g., $f_j$ and $f_k$ as shown in the figure. Based on such information, agent $i$ computes an initial importance score for itself and its neighbors (Section~\ref{sec:local_score}). Afterward, agent $i$ refines its initial score together with other agents in the system via a system-wide consensus algorithm, including first an update of its own score and then multiple iterations to reach system-wide consensus (Section~\ref{sec:consensus}). Note that during the consensus process, only scores are transmitted among events (not selected frames). After running the consensus algorithm, each agent will have the same copy of the final importance scores for their selected frames in the current period, defined as $\vec{x} = [x_1, x_2, ..., x_N]$. Agent $i$ then chooses its fast-forwarding strategy for the next period $s_i^{t+1}$ based on the rank of its importance score $x_i$ (Section~\ref{sec:strategy_selection}). More details are presented below and the notations are in Table~\ref{tab:notation}.
\begin{table}[htbp]
\begin{center}
\begin{tabular}{c|l}
\hline
$M$  & number of available fast-forwarding strategies\\
\hline
$N$  & number of camera views / agents\\
\hline
$V$ & the set of $N$ views $\left\{v_i\right\}$, $i \in [1, N]$  \\
\hline
$S$   & the set of available strategies $\left\{s^m\right\}$, $m \in [1, M]$ \\
\hline
$s_i^t$  & strategy being used in agent $i$ at adaptation step $t$\\
\hline
$s_i^{t+1}$   & strategy for agent $i$ in the next adaptation step $t+1$\\
\hline
$F$  & summary of the scene: $\left\{f_1,\cdots, f_N\right\}$ \\
\hline
$\vec{x}$ & importance score vector after consensus \\
\hline
$T$   & period of strategy update \\
\hline
\end{tabular}
\end{center}
\caption{Notation used in our proposed framework.}
\vspace{-12pt}
\label{tab:notation}
\end{table}

\subsection{Multi-strategy Fast-forwarding} 
\label{sec:agent}

On each camera that captures a view of the scene, we have a multi-strategy fast-forwarding agent that can adaptively fast-forward the incoming videos with different paces. In this work, we leverage a state-of-the-art fast-forwarding algorithm, FFNet~\cite{lan2018ffnet}, and derive three different strategies/paces for fast-forwarding: normal-pace, slow-pace, and fast-pace. Note that our approach can be easily extended to consider other numbers of strategies/paces.
\begin{figure}[t]
	\begin{center}
		\includegraphics[width=0.85\linewidth]{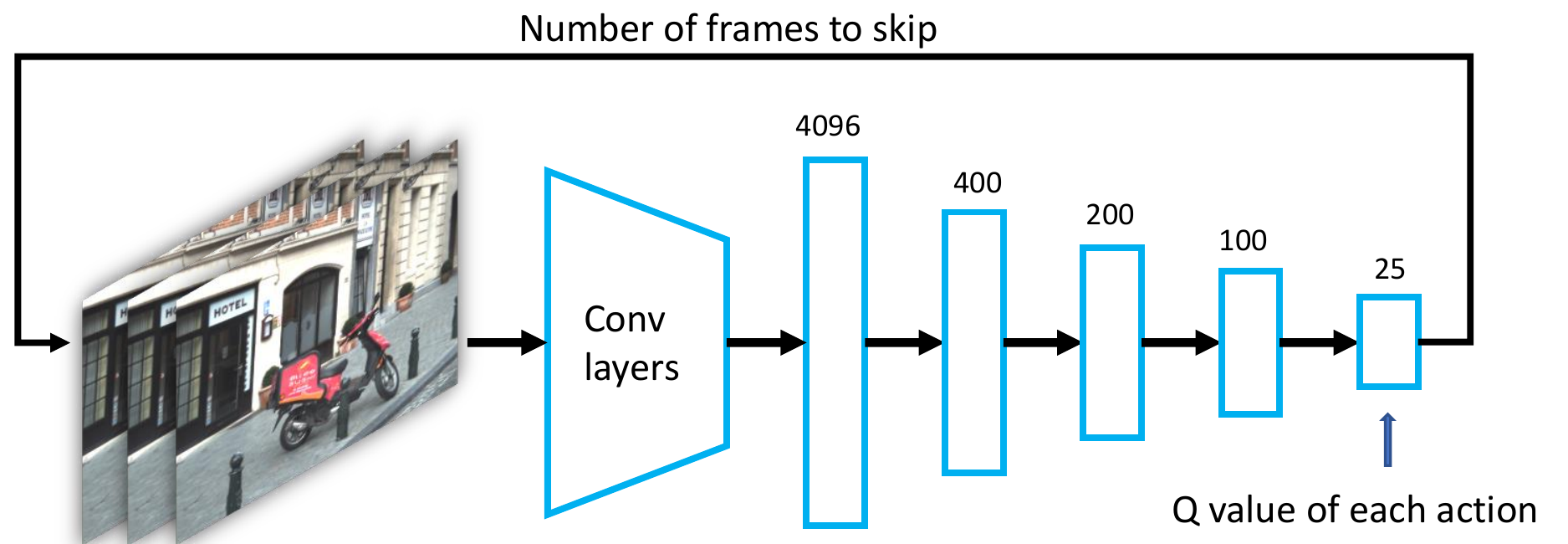}
	\end{center}
	\caption{\textbf{The model structure of normal-pace strategy.} It takes a frame in a video stream as an input for the deep neural network and outputs the number of frames to skip.}
	\vspace{-10pt}
	\label{ffnet}
\end{figure}

\medskip \noindent \textbf{Normal-pace Strategy:} The normal-pace strategy utilizes the same structure as FFNet~\cite{lan2018ffnet} (Fig.~\ref{ffnet}).
The fast-forwarding problem is modeled as a Markov decision process (MDP) and solved with a deep Q-learning (DQL) model. It learns a policy for skipping unimportant frames entirely and presenting the important ones for further processing. The state is defined as the feature of the current frame. The action set includes the possible numbers of frames to skip. An immediate reward at time step $k$ is defined as 
\begin{equation}
r_k(normal) =   - SP_k +HR_k,
\end{equation}
where $SP_k$ is the ``skip'' penalty and $HR_k$ is the ``hit'' reward for current action $a_k$. The skip penalty gives high scores for skipping unimportant frames and low scores for skipping important ones. The hit reward encourages the agent jumping to an important frame or a position near an important frame. With the definition of states, actions and rewards, a skipping policy is learned for selecting the action that maximizes the expected accumulated reward. As our normal-pace strategy, we use an action space of size 25, i.e., skipping from 1 to 25 frames.

\medskip \noindent \textbf{Slow-pace Strategy:} The slow-pace strategy aims at skipping fewer frames and thus retaining more frames in the selected buffer, possibly including more numbers of important frames. To meet this goal, we modify the immediate reward in FFNet at time step $k$ as 
\begin{equation}
r_k(slow) =   (- SP_k +HR_k) \times (1 - \frac{sigmoid(a_k)}{2}).
\end{equation}
Intuitively, if the agent skips with a larger step, it will receive smaller immediate reward. We also change the action space to 15 to prevent the agent from skipping too much.

\medskip \noindent \textbf{Fast-pace Strategy:} The goal of the fast-pace strategy is to skip more unimportant frames for more efficient processing and transmission. We modify the immediate reward at time step $k$ as
\begin{equation}
r_k(fast) =   (- SP_k +HR_k) \times (1 + \frac{sigmoid(a_k)}{2}).
\end{equation}
This reward definition ensures that the agent will get larger immediate reward if it skips with a larger step. The action space is set to 35 to allow the agents to skip larger steps.

In our framework, each agent updates its fast-forwarding strategy periodically based on the evaluation of the relative importance of its selected frames in the current period (when compared with other agents). Lower importance may lead to a faster strategy for the next period for reducing data processing and transmission, while higher importance may lead to a slower strategy for reducing the likelihood of missing important frames.

\subsection{Local-neighbor Importance Score}
\label{sec:local_score}

In this step, for every agent $i$, we compute an initial importance score for itself and its neighbors by comparing the similarities between their selected frames.
First, we evaluate the similarity between two frames $x$ and $y$ by computing the exponential of the scaled negative L2-norm of the feature representations of the two frames, as defined in the following equation.
\begin{equation}
    sim(x, y) = e^{-\alpha||x-y||_2},
\end{equation}
where $\alpha$ is used to scale the L2-norm to restrict the similarity value to a satisfactory range. In the experiment, we set $\alpha = 0.05$.

The similarity of agent $j$ to $i$ is then defined as 
\begin{equation}
    sim\_agent (v_i, v_j) = \frac{1}{|v_j|} \sum_{s=1}^{|v_j|} \max_{1\leq a \leq |v_i|} sim(p_s(v_j), p_a(v_i)),
\end{equation}
where $|v_j|$ denotes the number of selected frames from agent $j$ and $p_s(v_j)$ denotes a selected frame $s$ from agent $j$. The similarity for frame $p_s(v_j)$ to agent $i$ is the maximum among the similarities between $p_s(v_j)$ and frames of agent $v_i$. Then the agent-to-agent similarity of agent $j$ to agent $i$ is the average frame similarity.  

We define the communication connections among agents as an undirected graph $G=(V,E)$. With this definition, we compute the importance score of agent $j$ estimated by agent $i$ as 
\begin{equation}\label{equ:xij0}
    x_{ij}^0 = \begin{cases}
        \frac{1}{|V_i|-1} \sum_{v_k \in V_i, k \neq j} sim\_agent (v_j, v_k) & \text{if } i=j \text{ or } (i,j) \in E  \\ 0  & \text{o.w.} 
\end{cases}
\end{equation}
where $V_i= \left\{v_k | (i,k) \in E\right\} \bigcup \left\{v_i\right\}$, is the set of the neighbors of agent $i$ and itself. $|V_i|$ represents the number of agents in $V_i$. This initial important score will then be refined via a consensus process.

\subsection{System-wide Importance Score Consensus} \label{sec:consensus}

To refine the initial importance score and reach an agreement across all agents on the relative importance of their frames, we mainly use a maximal consensus algorithm in our framework. We have also explored multiple variants of our framework with different consensus methods. 

\medskip \noindent \textbf{Maximal Consensus Algorithm in DMVF:} There are three steps in this algorithm. First, each agent communicates with its neighbors and sends its initial importance scores for each of them. At the end of this step, agent $i$ will have the initial scores of itself from its own computation and from the evaluation by its neighbors (i.e. \{$x_{ji}^0$ \}, $j \in V_i$). Then, in the second step, agent $i$ updates its score as %
    \begin{equation}\label{equ:importantscore}
        x_i = \frac{\sum_{j \in V_i} \frac{1}{n_j} x_{ji}^0}{\sum_{j \in V_i} \frac{1}{n_j}},
    \end{equation}
which means the importance score of agent $i$ is updated as the weighted average of the initial importance scores evaluated by itself and its neighbors. Then an importance score vector $\vec{x}_i$ for all agents is constructed by agent $i$, with only the $i$-${th}$ element set to $x_i$ and all others set to zero. In the third step, all agents will run a maximal consensus algorithm over the importance score vector. This algorithm only requires the number of consensus steps to be the diameter of the graph $G$ to reach an agreement (the convergence is guaranteed). In the end, every agent will have the same copy of the importance score vector for all agents, i.e., $\vec{x}_i =\vec{x} = [x_1, x_2, ..., x_N]$. Details of the importance score consensus is shown in Algorithm~\ref{alg1}. 

\begin{algorithm}[t]
	\caption{Algorithm for Computing Importance Score}
	\label{alg1}
	\begin{algorithmic}[1]
    	\renewcommand{\algorithmicrequire}{\textbf{Input:}}
        \renewcommand{\algorithmicensure}{\textbf{Output:}}
 		\REQUIRE Selected frames from agents at current period: $f_1, f_2, ..., f_N$ 
		\ENSURE List of important scores $\vec{x}_i$ for each agent/view $v_i\in V$
		\FOR{each agent $v_i\in V$}
		\STATE Send selected frames $f_i$ to neighbors.
		\STATE $V_i = \{v_i$'s neighboring agents and itself$\}$.
		\FORALL{$v_j\in V_i$}
		\STATE $f_j \leftarrow Receive\_Selected\_Frames\_From(v_j)$.
		\ENDFOR
		\FORALL{$v_j\in V_i$}
		\STATE $x_{ij}^0 = \frac{1}{|V_i|-1} \sum_{v_k \in V_i, k \neq j} sim\_agent (v_j, v_k) $.
		\STATE Send $x_{ij}^0$ to $v_j$.
		\ENDFOR
		\FORALL{$v_j\in V_i$}
		\STATE $x_{ji}^0 \leftarrow Receive\_Initial\_Score\_From(v_j)$.
		\ENDFOR
		\STATE $x_i = \frac{\sum_{j \in V_i} \frac{1}{n_j} x_{ji}^0}{\sum_{j \in V_i} \frac{1}{n_j}}$.
		\STATE $\vec{x}_i = [0,0,\ldots, 0]$, $\vec{x}_i[i] = x_i$.
		\FOR{$t = 1:graphDiameter$}
		\STATE Send $\vec{x}_i$ to neighbors.
		\FORALL{$v_j\in V_i$}
		\STATE $\vec{x}_j \leftarrow Receive\_Score\_Vector\_From(v_j)$.
		\FOR{$k = 1:N$}
		\STATE $\vec{x}_i[k] = \max(\vec{x}_i[k], \vec{x}_j[k])$
		\ENDFOR 
		\ENDFOR
		\ENDFOR
		\RETURN $\vec{x}_i$
		\ENDFOR
	\end{algorithmic}	
\end{algorithm}

\medskip \noindent \textbf{Other Consensus Methods:} Aside from the main DMVF with maximal consensus algorithm, we also developed multiple variants of our framework with different consensus methods, including DMVF-DGD, DMVF-EXTRA, DMVF-AVE, and DMVF-ONE.
\begin{myitemize}
    \item \textbf{ DMVF-DGD.} In this design, we leverage the distributed gradient descent (DGD)~\cite{nedic2009distributed} method for reaching consensus on the importance score of every agent. Each consensus step can be represented by multiplication by an $N\times N$ row stochastic consensus matrix $P$, with $P_{ij}\neq 0$ if and only if $(i,j)$ in $E$. 
    In the experiment, the consensus matrix $P$ is defined as 
    \begin{equation}
        \label{eqn:p}
        \begin{aligned}
        P_{ij} &= \frac{1}{d_{max}+1}\quad & i\neq j, (i,j)\in E,\\
        P_{ij} &= 0 \quad & i\neq j, (i,j)\notin E,\\
        P_{ii} &= \frac{d_{max}+1-d_i}{d_{max}+1}\quad & i = j,
        \end{aligned}
    \end{equation}
    where $d_i=n_i$ is the degree of agent i (i.e., number of neighbors of agent i), $d_{max}=\max_i\{d_i\}$ is the maximum degree in the system. The objective of this distributed optimization is
    \begin{equation}
    \label{eqn:obj}
        \sum_i f_i(\vec{x}) = \sum_i \left( \frac{1}{n_i} \sum_{(i,j) \in E} (x_{j}- x_{ij}^0)^2 \right)
    \end{equation}
    where $\vec{x}$ is the importance scores for all agents and $x_j$ is the $j$-${th}$ element corresponding to the importance score of agent $j$. $x_{ij}^0$ is agent $j$'s score evaluated by agent $i$ (computed by Equation~\eqref{equ:xij0}). $n_i$ is the total number of neighbors of agent $i$.
    The DGD iteration step is as follows:
        \begin{equation}
            \vec{x}_i^{t+1} = \sum_{j} P_{ij} \vec{x}_j^t - \gamma^t \nabla f_i (\vec{x}_i^t)
        \end{equation}
    where $P_{ij}$ is the $(i,j)$ element of the consensus matrix $P$ and the stepsize $\gamma^t$ is gradually reduced based on $\gamma^t \sim \frac{1}{t}$.
        
    \item \textbf{DMVF-EXTRA:} In this variant, we utilize the decentralized exact first-order algorithm (EXTRA)~\cite{shi2015extra} as the consensus method in the framework. EXTRA has the same consensus matrix and objective as DGD in Equation~\eqref{eqn:p} and Equation~\eqref{eqn:obj}, respectively. The EXTRA iteration step is as follows:
    \begin{equation}
        \vec{x}^{t+2}_i = \sum_j M_{ij} \vec{x}^{t+1}_j - \sum_j \frac{M_{ij}}{2} \vec{x}^{t}_j - \alpha [\nabla f_i(\vec{x}^{t+1}_i) - \nabla f_i(\vec{x}^t_i)]
    \end{equation}
    where $M = I+P$ and $M_{ij}$ is the $(i,j)$ element of $M$. $\alpha$ is a constant for the iteration stepsize.
    \item \textbf{DMVF-AVE:} First, each agent sends its initial importance score evaluations of its neighbors to them. Then, every agent $i$ updates its score by taking the average of the initial score evaluations of itself from its own computation and from its neighbors:
    \begin{equation}
        x_i = \frac{\sum_{j \in V_i } x_{ji}^0 }{n_i + 1}.
    \end{equation}
    Then, similar to the consensus algorithm in DMVF, all agents will run a maximal consensus algorithm on the importance scores.
    
    \item \textbf{DMVF-ONE:} In this design, each agent takes $x_{ii}^0$ as its updated score, i.e., $x_i = x_{ii}^0$. Then, all agents will run a maximal consensus algorithm on the importance score.  
\end{myitemize}

\subsection{Strategy Selection}
\label{sec:strategy_selection}

Based on the final importance scores in $\vec{x}$, the agents with higher scores could be assigned with a slower strategy for the next period, while the agents with lower scores could be faster.
In our framework, given the system requirement, the portion of different strategies are pre-defined, which means there should be a fixed number of agents under each strategy after every update. In our experiment, there are three strategies for 6 agents. The system requirement is denoted as $X/Y/Z$, $X+Y+Z = 6$, which means the system requires $X$ agents to use the fast strategy, $Y$ agents to use the normal strategy and $Z$ agents to use the slow strategy. The strategy for each agent will then be assigned based on the ranking of its importance score in $\vec{x}$.

\section{Experimental Results}

In this section, we present the experimental results of our DMVF framework and its comparison with several methods in the literature. We also demonstrate the trade-off between coverage and efficiency in DMVF, evaluate how various degrees of communication network connectivity affect performance of DMVF, present analysis on different consensus methods, and report timing efficiency. The source code can be found at \href{https://github.com/shuyueL/DMVF}{https://github.com/shuyueL/DMVF}.

\subsection{Experimental Setup}

\noindent \textbf{Dataset:} We evaluate the performance of our framework on a publicly available multi-view video dataset VideoWeb~\cite{denina2011videoweb}. It consists of realistic scenarios in a multi-camera network environment that involves multiple persons performing dozens of different repetitive and non-repetitive activities. We use the Day 4 subset of the dataset, as it involves multiple vehicles and persons. It has 6 scenes and each scene has 6 views of videos. All videos are captured at $640 \times 480$ resolution and approximately 30 frames/second. We transfer the given labels of important actions to binary indicators of important frames. A global ground truth that combines all important intervals across views is generated for evaluating the performance. 

We also looked into other multi-view datasets, such as the well-known Office, Campus, Lobby, Road, Badminton datasets from~\cite{fu2010multi}, and BL-7F from~\cite{ou2015line}. The first class of datasets are all of small sizes, with only 3-4 videos (1 video per view) available for each scene. The same problem occurs in BL-7F dataset as it does not have enough overlapping videos for training of the fast-forwarding agent.

\smallskip
\noindent \textbf{Evaluation Metrics:} 
Similar to~\cite{lan2018ffnet}, we consider a \emph{coverage} metric at frame level, which evaluates how well the results from the fast-forwarding methods cover the important frames labeled in the ground truth. The coverage is computed as the percentage of covered frames by all agents in the global ground truth. If an important frame is covered by any of the agents, it will be considered as true positive.
We also evaluate the efficiency of various methods by considering their \emph{processing rate}, i.e., the average percentage of frames processed at the agents. The processing rate measures the computation load of the agents. In this work, the communication load can also be indicated by the processing rate, since the number of transmitted frames is proportional to the number of processed frames (with the addition of processed frames from the neighbors). 

\smallskip \noindent \textbf{Implementation Details:} The multi-strategy fast-forward agents are implemented using the TensorFlow library and modeled as 4-layer neural networks. $\epsilon$-greedy strategy is used to better explore the state space during the training process. The strategy update period $T$ is set to 100 frames of the raw video inputs.  The three strategies used in our framework are defined in Section~\ref{sec:agent}. The operating points of agents with the slow, normal and fast strategies are shown in Table~\ref{tab:per_stra}. 
\begin{table}[ht]
\begin{center}
\begin{tabular}{c|c|c|c}
\hline
Strategy & Slow & Normal & Fast \\
\hline
Processing rate(\%) & 8.69 & 6.02 & 3.73\\
\hline
Coverage(\%) & 73.45 & 61.91 & 55.89 \\
\hline
\end{tabular}
\end{center}
\caption{Operating points of 3 fast-forwarding strategies.}
\vspace{-12pt}
\label{tab:per_stra}
\end{table}

\begin{figure}[t]
	\begin{center}
		\includegraphics[width=0.85\linewidth]{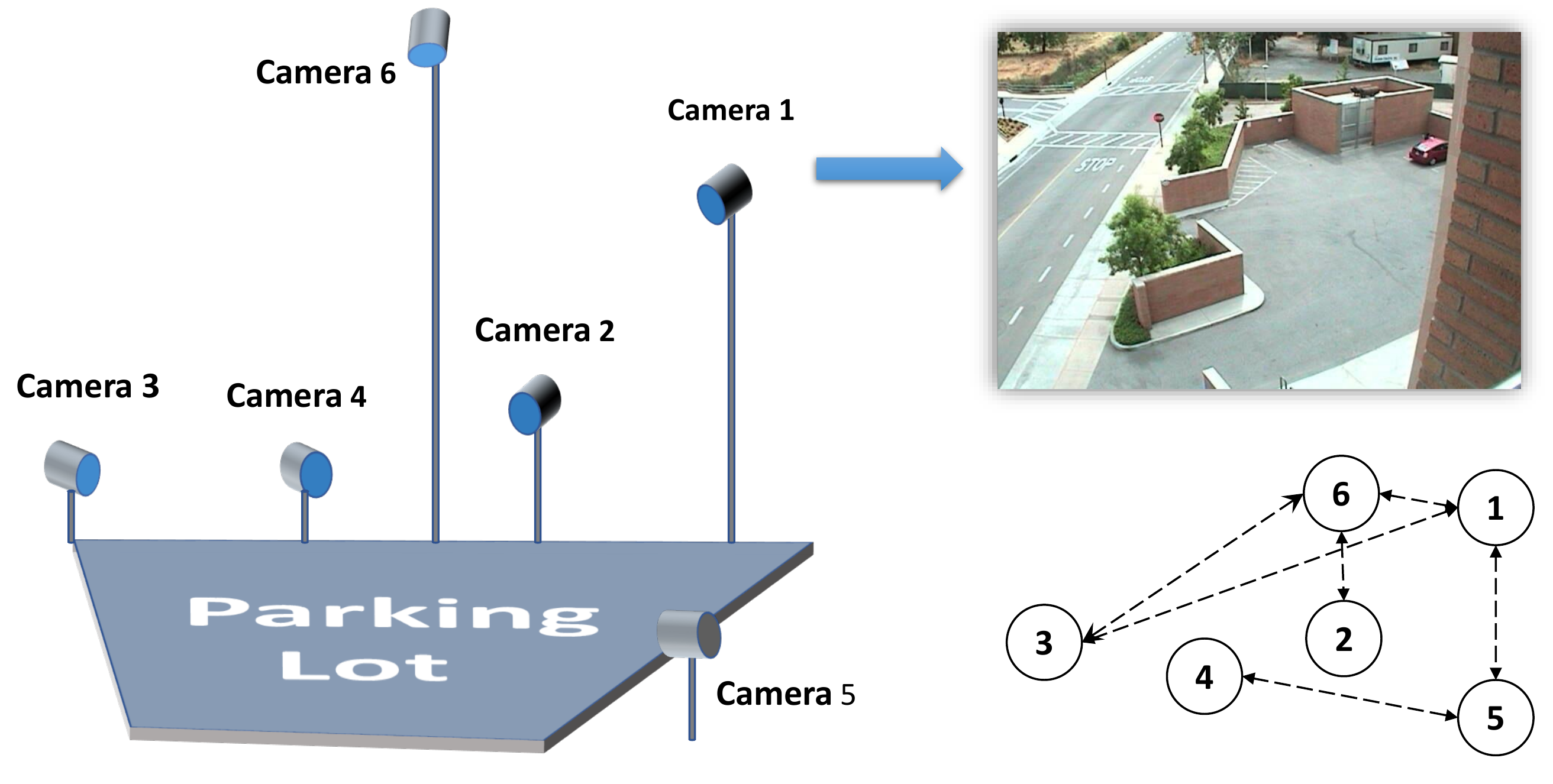}
	\end{center}
	\caption{\textbf{The communication graph of agents.} The left part of the figure shows the physical locations of the cameras and the environment. The right-bottom graph is the communication graph of the agents for the main set of experiments (built according to the similarity of views).}
	\vspace{-10pt}
	\label{graph}
\end{figure}

In our main set of experiments (except for the study in Section~\ref{sec:exp_connection}), we evaluate the proposed framework on the aforementioned dataset with a communication graph of agents based on their view similarity, as shown in Fig.~\ref{graph}. We deploy the proposed DMVF on an actual embedded platform. Five agents are implemented on 2 workstations and 3 laptops, and the other one is run on an Nvidia Jetson TX2. The communication between agents is implemented with WiFi network using TCP protocol.

\smallskip \noindent \textbf{Comparison Methods:}
We compare our approach with several methods for video fast-forwarding and video summarization, which include both online and offline methods. 
(1) \textbf{FFNet}~\cite{lan2018ffnet}. FFNet is an online single-agent fast-forwarding approach that utilizes Q-learning. Applying FFNet to multi-agent scenarios assumes that all agents perform video fast-forwarding independently without any communication between them.
(2) \textbf{Random}. The random method skips the incoming frames randomly. 
(3) \textbf{Uniform}. The uniform method skips the incoming frames periodically.
(4) \textbf{OK} (Online Kmeans)~\cite{arthur2007k}. Online Kmeans is a classical clustering based method working in an online update fashion.  The frames that are the closest to the centroid in each cluster are selected as the summary.
(5) \textbf{SC} (Spectral Clustering)~\cite{von2007tutorial}. Spectral Clustering is a different clustering based method to group all the frames in a video to several clusters. The summary is composed by the frames that are closest to each centroid.
(6) \textbf{SMRS} (Sparse Modeling Representative Selection)~\cite{elhamifar2012see}. This is an offline method that requires all videos available before the processing and outputs the summary of each video. It takes the entire video as the dictionary and finds the representative frames based on the zero patterns of the sparse coding vector. 

\begin{table*}[t]
\begin{center}
\begin{tabular}{c|c|c|c|c|c|c|c}
\hline
Methods & Random & Uniform & OK & SC & SMRS & FFNet & DMVF \\
\hline
Coverage(\%) & 50.78 & 25.80 & 50.21 & 44.74 & 42.36 & 61.91 &  65.87\\
\hline
Processing rate(\%) & 4.20 & 3.70 & 100 & 100 & 100 & 6.02 &  5.06\\
\hline
\end{tabular}
\end{center}
\caption{\textbf{Comparison on coverage and processing rate between DMVF and other approaches.} Compared with FFNet, DMVF achieves better coverage (6.40\% improvement) while reducing the processing rate by 15.95\%. For other methods, DMVF achieves either much better coverage or much lower processing rate or both. }
\vspace{-10pt}
\label{tab:main}
\end{table*}

\smallskip \noindent \textbf{Experimental Settings:}
We use the penultimate layer (pool 5) of the GoogLeNet
model~\cite{szegedy2015going} (1024-dimensions) to represent each video frame. For each method, we tune its parameters for best performance, and control them to keep the fast-forwarded videos to the same percentage of raw frames for a fair comparison. During evaluation, we extend the frames selected by each method before and after by a small window, whose size is the same for all methods. For FFNet, we keep all the settings as described in their paper and train it on the VideoWeb dataset. For OK and SC, we set the number of clusters to 20. We randomly use 80\% of the videos for training and the remaining 20\% for testing. 5 rounds of experiments are run and the reported result is the average performance.

\subsection{Comparison with Other Approaches}

Table~\ref{tab:main} shows the coverage metric and the processing rate of DMVF on the VideoWeb dataset and its comparison with other approaches. In this experiment, the system requirement is 3/2/1, i.e., 3 fast strategies, 2 normal strategies and 1 slow strategy for the agents (further study on different system requirements is shown later). 
From the table, we can clearly see the improvement from DMVF:
\begin{myitemize}
    \item Compared with the state-of-the art method for fast-forwarding, FFNet, our approach DMVF achieves better coverage (6.40\% improvement) while reduces the processing rate by 15.95\%. 
    \item For those methods that require processing the entire video (processing rate of 100\%), i.e., OK, SC and SMRS, our framework DMVF achieves higher coverage (more than 25\% increase) and much lower processing rate.
    \item Compared with Random and Uniform methods, DMVF offers significant improvement in coverage with modest increase of processing rate.
\end{myitemize}

\begin{figure}
\begin{center}
     \includegraphics[width=0.9\linewidth]{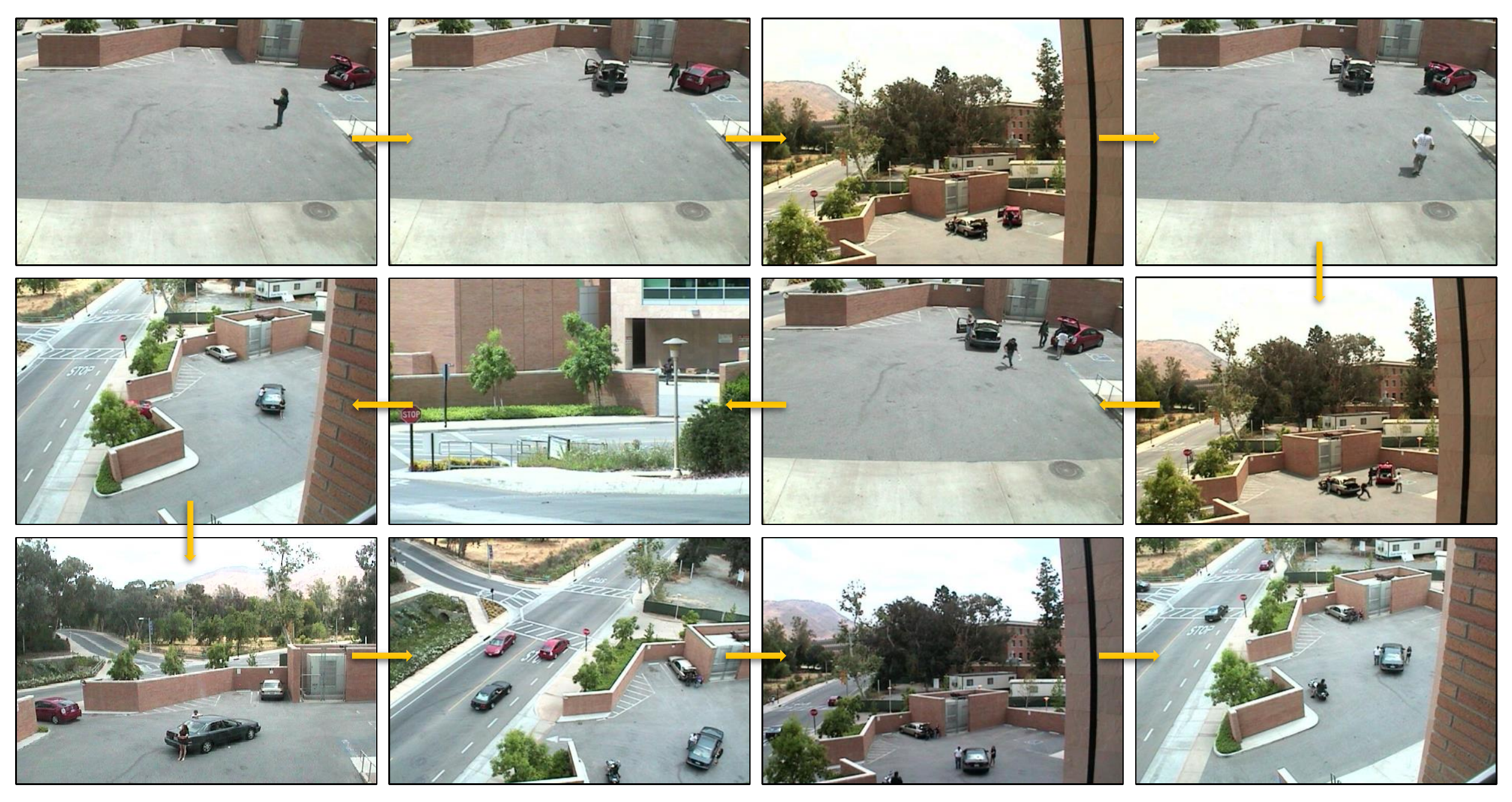}
\end{center}
  \caption{\textbf{Representative frames generated by DMVF from the VideoWeb dataset.}}
\label{qualitative}
\vspace{-12pt}
\end{figure}

Fig.~\ref{qualitative} shows a qualitative example for fast-forwarding videos with DMVF. As we can see, it selects more important frames from multiple views, and creates a compact subset of frames.

\subsection{Coverage-Efficiency Trade-off}

\begin{figure}[t]
	\begin{center}
		\includegraphics[width=0.9\linewidth]{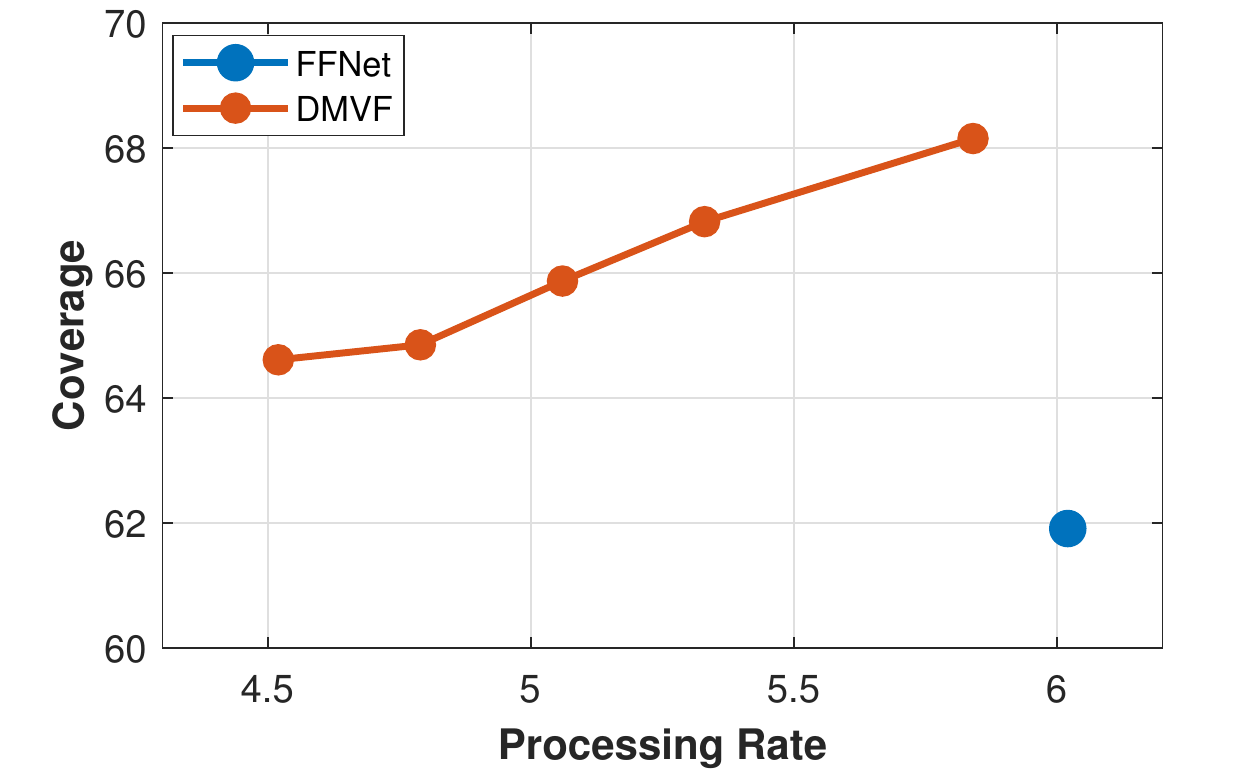}
	\end{center}
	\caption{Trade-off between coverage and processing rate in DMVF, under different system requirements.}
	\vspace{-10pt}
	\label{tradeoff}
\end{figure}

Table~\ref{tab:main} shows that DMVF can significantly outperform other approaches in the literature.  Furthermore, as shown in this section, DMVF enables flexible coverage-efficiency trade-off with different system requirements. Note that when deploying a video fast-forwarding strategy, the goal of achieving high efficiency (i.e., low processing rate) contradicts the goal of maintaining high coverage, and the ability to trade off between the two is desirable.

Fig.~\ref{tradeoff} shows that different trade-offs between coverage and efficiency can be easily achieved in DMVF by changing the X/Y/Z parameters in system requirement. The points on the red curve of DMVF are achieved by the following X/Y/Z: 2/2/2, 2/3/1, 3/2/1, 4/1/1, and 5/0/1. All points outperform the FFNet result. Note that changing these parameters is much more flexible and systematic than deploying FFNet on each agent and manually adjusting their skipping speeds, showing DMVF's capability for reconfiguration and adaptation to accommodate varying system operation needs.

\subsection{Impact of Connectivity}
\label{sec:exp_connection}

\begin{figure}[t]
    \centering
    \begin{subfigure}[t]{\columnwidth}
        \centering
        \includegraphics[width=0.9\linewidth]{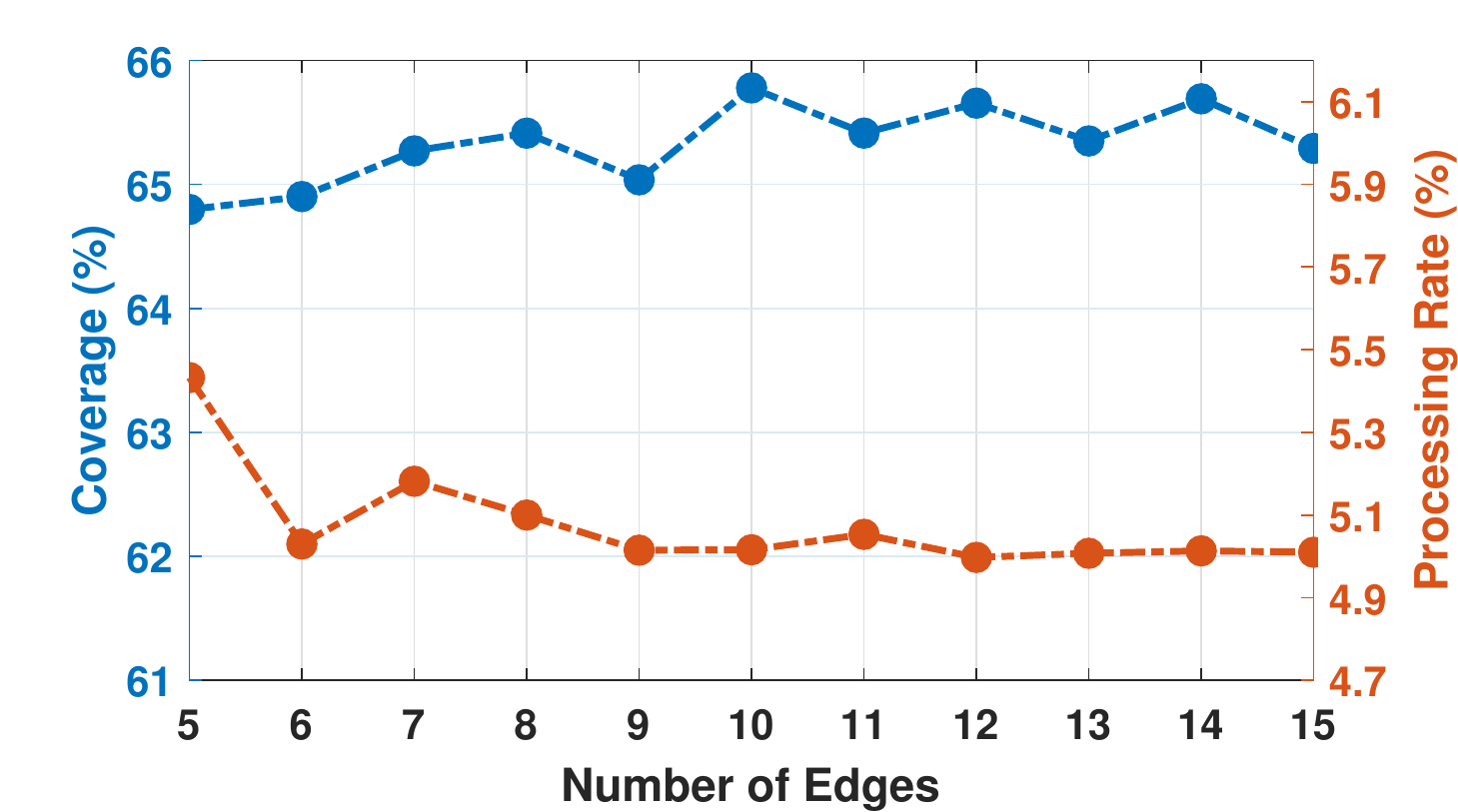}
        \caption{  }
    \end{subfigure}
    \begin{subfigure}[t]{\columnwidth}
        \centering
        \includegraphics[width=0.9\linewidth]{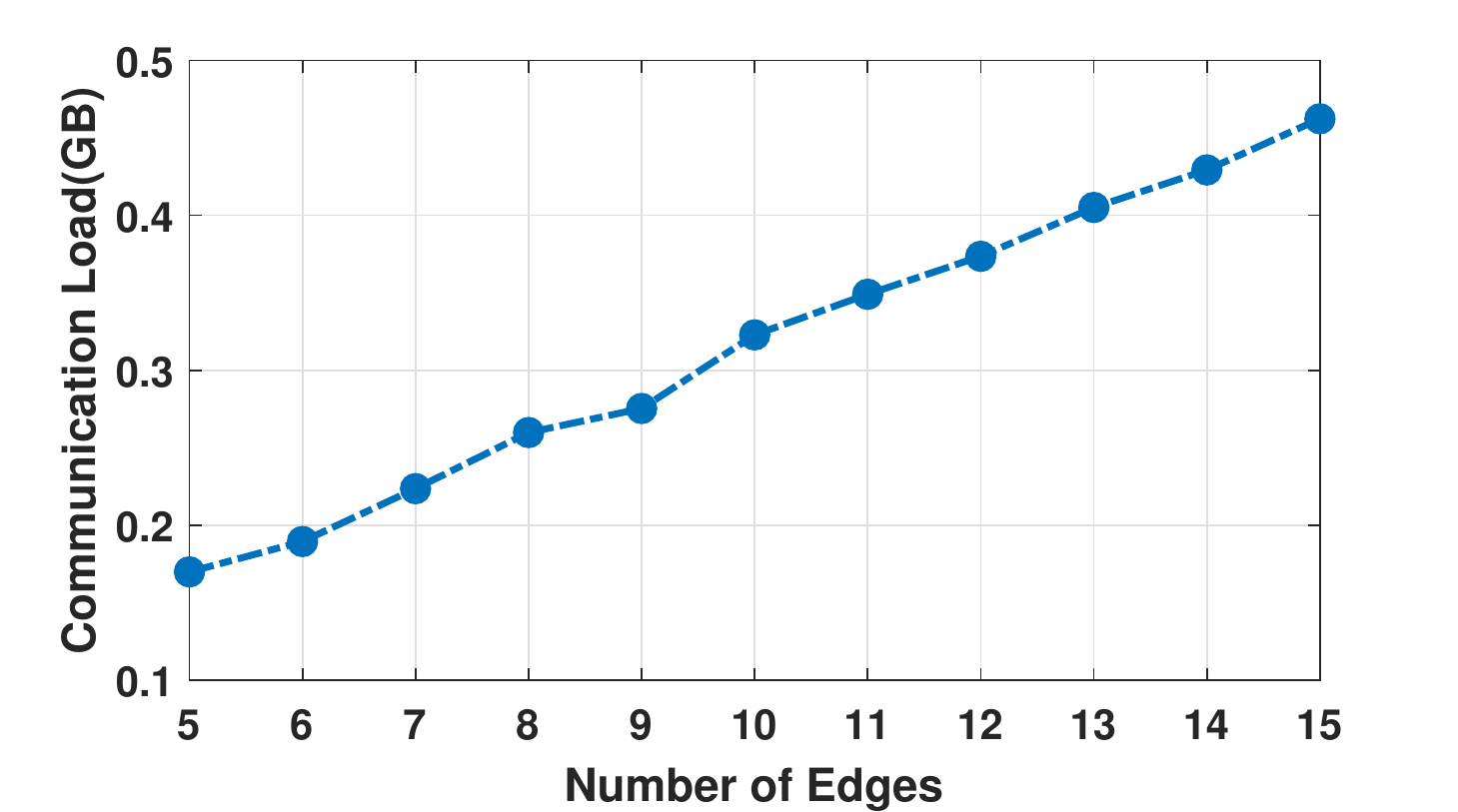}
        \caption{ }
    \end{subfigure}
    \caption{Coverage, processing rate and communication load under varying degree of connectivity in DMVF.}
    \label{comm}
%   \vspace{-12pt}
\end{figure}

For a distributed multi-agent application, such as DMVF, connection pattern among agents and its resulting communication load is an important factor. In real applications, the connections among different agents may vary because of the connection capacity of agents, physical distance between agents, and the network bandwidth, etc. Here, we evaluate the performance and communication load of the DMVF framework under wireless communication with different connection patterns. 
We use Erdos-Renyi method to generate random graphs as the connection patterns among agents, as shown in Fig.~\ref{comm}. Each random graph is generated with a parameter $P$, i.e., a fixed probability of each edge being present or absent, independent of the other edges. Higher probability yields higher connectivity of the multi-agent systems. We generate over 40 graphs with different values of $P$ and evaluate their performance with respect to different number of edges in the graph (results are averaged for each number of edges). 
In the experiment, the total raw input data is 12.36 GB. We observe the following from the results.
\begin{myitemize}
    \item From Fig.~\ref{comm} (a), we can observe the robustness of DMVF under varying degrees of connectivity. DMVF maintains a higher coverage and lower processing rate when compared to FFNet, regardless of the number of edges in the connection graph.
    \item From Fig.~\ref{comm} (b),   
    DMVF leads to only a small amount of communication overhead, from 0.17 GB to 0.46 GB under full connectivity (1.37\% to 3.72\% of raw data).
\end{myitemize}

\subsection{Analysis on Consensus Algorithms}

\begin{figure}[t]
	\begin{center}
		\includegraphics[width=0.9\linewidth]{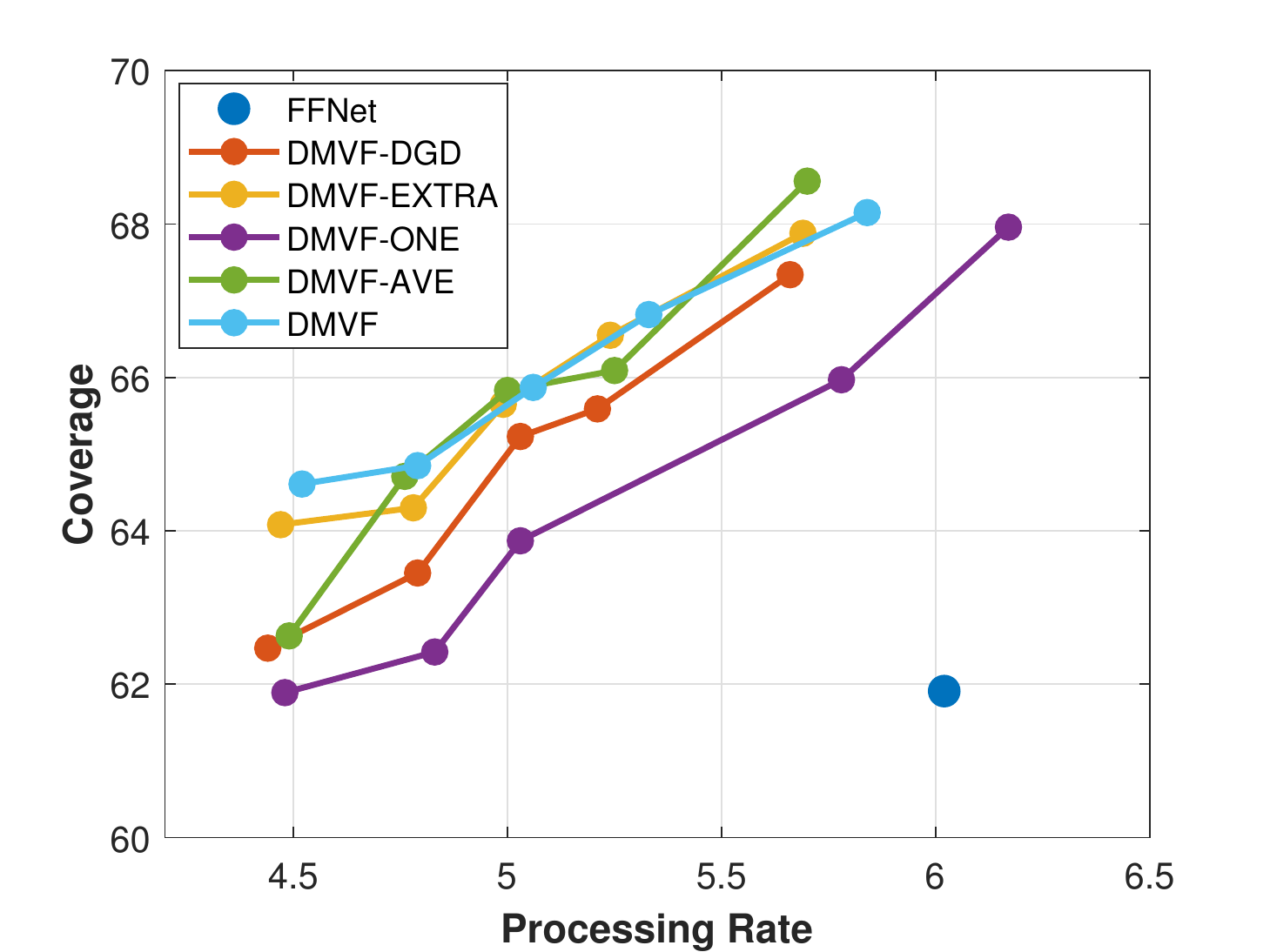}
	\end{center}
	\caption{Comparison between DMVF with different consensus algorithms and FFNet.}
	\label{abla_sim}
\end{figure}

Here, we evaluate DMVF with the various consensus algorithms introduced in Section~\ref{sec:consensus} (communication graph is as in Fig.~\ref{graph}). Fig.~\ref{abla_sim} shows the comparison between FFNet and DMVF variants with respect to average coverage and processing rate. We also evaluate the number of iterations each method needed to reach consensus, and list them together with the data from Fig.~\ref{abla_sim} in Table~\ref{tab:cov_proc}. From Fig.~\ref{abla_sim} and Table~\ref{tab:cov_proc}, we have the following observations:
\begin{myitemize}
    \item Compared to FFNet, any variant of DMVF outperforms both in coverage and in processing rate.
    \item Compared to the gradient descent based variants, i.e., DMVF-DGD and DMVF-EXTRA, the proposed DMVF needs much fewer iterations to reach consensus.
    \item Among the three maximal consensus based variants, i.e. DMVF, DMVF-AVE, and DMVF-ONE, DMVF has the highest average coverage while having the same magnitude of iteration count. 
\end{myitemize}

\begin{table}[t]
\begin{center}
\begin{tabular}{c|c|c|c}
\hline
Methods & Ave. Coverage & Ave. Proc. Rate & Iterations\\ 
\hline
FFNet & 61.91\% & 6.02\% & \# \\
\hline
DMVF-AVE & 65.56\% & 5.04\% & 5\\
\hline 
DMVF-ONE & 64.42\% & 5.26\% & \textbf{4}\\
\hline
DMVF-EXTRA & 65.69\% & \textbf{5.03\%} & 93.0\\
\hline 
DMVF-DGD & 64.82\% & \textbf{5.03\%} & 234.48\\
\hline
DMVF & \textbf{66.06\%} & 5.11\% & 5\\
\hline
\end{tabular}
\end{center}
\caption{Comparison of coverage, processing rate and iterations of DMVF with different consensus methods.}
% \vspace{-12pt}

\label{tab:cov_proc}
\end{table}

\subsection{Timing Efficiency}
We evaluate the timing efficiency of DMVF on the aforementioned distributed embedded platform. 
Our framework achieves an average frame rate of 313 FPS and a worst-case rate of 280 FPS. 
Such high frame rate may not be achieved on less-capable embedded platforms, but the low processing rate from DMVF should still reduce computation cost and achieve near real-time speed. For example, even when we intentionally slow down the computation by only using the ARM cores on TX2 (i.e., not using the GPU), DMVF achieves 94 FPS, which is sufficient for most real-time monitoring cases.

\section{Conclusion}
In this paper, we propose a distributed multi-agent video fast-forwarding framework, aka DMVF, that optimizes the coverage and processing rate by enabling agents associated with different cameras to communicate with their neighbors and adaptively update their reinforcement learning based fast-forwarding strategies. Our experiment shows the effectiveness of DMVF when compared with other fast-forwarding methods in the literature. 
% The codes are publicly available at: \href{https://github.com/shuyueL/DMVF}{https://github.com/shuyueL/DMVF}.

%\clearpage

\begin{acks}
We gratefully acknowledge the support from NSF grants 1834701, 1834324, 1839511, 1724341, and ONR grant N00014-19-1-2496. Roy-Chowdhury also acknowledges support from CISCO.
\end{acks}

\pagebreak
\bibliographystyle{ACM-Reference-Format}
\bibliography{shuyue_bib}

\end{document}